\documentclass[letterpaper, 10 pt, conference]{ieeeconf}  

\IEEEoverridecommandlockouts                              

\usepackage{graphics} 
\usepackage{bm}
\usepackage{dblfloatfix}
\usepackage{cuted}
\usepackage{epsfig} 
\usepackage{mathptmx} 
\usepackage{times} 
\usepackage[T1]{fontenc} 
\usepackage{amsmath} 
\usepackage{amssymb}  
\usepackage{cite}     
\newcommand{\GOLEM}{\textsc{Golem}}
\usepackage{booktabs}
\usepackage{graphicx}
\usepackage{xcolor}
\usepackage{caption}
\usepackage{amsmath,bm}
\usepackage{subcaption}
\usepackage{url}
\usepackage{hyperref}
\usepackage{xcolor}

\newcommand{\na}{\textcolor{black!40}{---}}      

\title{\LARGE \bf
\GOLEM: Modular Humanoid Autonomy Towards Electric Vehicle Battery Disassembly
}

\author{
Max Conway$^{1*}$,
William Xie$^1$,
Allen Devaraj$^1$,
Yutong Zhang$^1$,
Niraj Pudasaini$^2$,
Mateo Feit,
Adam Abid$^1$,
\\
Zachary Allen$^1$,
Chen Liu$^2$,
Xuan Tan$^1$,
Jensen Lavering$^1$,
Jason Chen$^1$,
Lyle Antieau$^1$,
\\
Anthony Von Pischke, %
Alessandro Roncone$^1$,
Zachary Sunberg$^1$,
Nikolaus Correll$^{1,2}$
\thanks{$^1$Department of Computer Science, University of Colorado Boulder, USA.}
\thanks{$^2$Department of Aerospace and Mechanical Engineering, University of Notre Dame, USA.}
\thanks{$^*$Corresponding author.}
}

\begin{document}

\maketitle
\thispagestyle{empty}
\pagestyle{empty}


\begin{abstract}
Disassembling end-of-life electric vehicle (EV) battery packs is dull and dangerous work, performed almost entirely by humans. We present GOLEM (Generalized Open Library of Embodied Modules), an end-to-end, open-source system architecture for EV battery disassembly with the Unitree H1-2 humanoid robot in which walking, manipulation, dynamic stability, navigation, and spatial memory are independent modules with abstract interfaces, so that methods are easily developed, interchanged, and compared. GOLEM is deployed as a Docker-based ROS 2 abstraction in which MuJoCo and IsaacLab digital twins expose interfaces matching the physical robot. GOLEM's composability and per-module customization enable development and demonstration of humanoid EV battery disassembly, from simulation to reality. GOLEM provides fair comparison between humanoid modules, enabling evaluation as a capability ladder, in which one module is characterized at a time and added as a rung: LiDAR-inertial navigation places the robot within 13.0cm of a 6m goal; a learned standing controller recovers from external disturbances that sampling-based lower-body MPC does not; and grasping loosened fasteners from a real Hyundai Ioniq 5 pack degrades from 97\% tethered to 87\% free-standing to 37\% under navigation-induced pose variance. 
Source code is available at the project page \url{https://golem-humanoid.github.io}.
\end{abstract}

\vspace{-6pt}
\section{Introduction}
\vspace{-2pt}

End-of-life electric vehicle lithium-ion batteries (EV-LIBs) must be disassembled so that precious minerals and metals therein can be recovered for reuse \cite{harper2019recycling}. This work is repetitive and hazardous: hundreds of fasteners must be removed from energy-dense packs that arrive in an unknown state of charge, can combust, emit toxic fumes, or electrocute workers \cite{allen2026robotic}. 

Dedicated automation \cite{wegener2015robot,choux2021task,allen2026robotic} is not versatile enough for the large variety of pack designs. Furthermore, recycling facilities are currently designed around humans positioning humanoid robots, which integrate without special accommodations such as ramps or fixed mounts, as appealing candidates. 
Proving humanoid efficacy in such hazardous industrial tasks represents a step toward more sensitive domains such as the home, healthcare, or disaster response.
\begin{figure}[!h]
    \centering
    \includegraphics[width=0.71\linewidth]{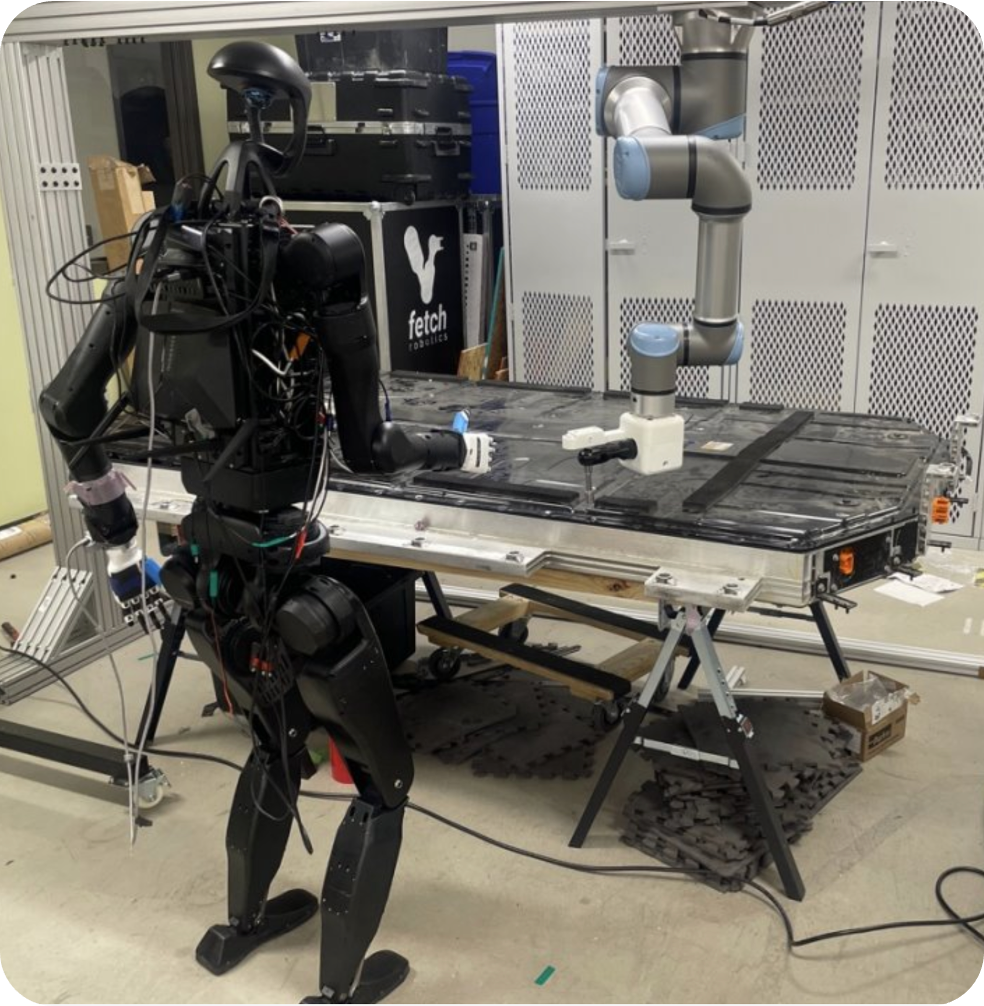}
    \caption{A Unitree H1-2 humanoid working alongside the gantry-based RAPID disassembly cell \cite{allen2026robotic} on a Hyundai Ioniq 5 battery pack. \GOLEM~augments automation infrastructure designed around human workers with humanoid autonomy.}
    \label{fig:teaser}
    \vspace{-20pt}
\end{figure}
While recent advances in learned whole-body control \cite{luo2025sonic, kim2026legs, shi2025almi} have significantly expanded humanoid capabilities, existing end-to-end paradigms remain ill-suited for industrial EV disassembly. 
First, these policies primarily target smaller platforms such as the Unitree G1 (1.32\,m, 35\,kg, 120\,Nm knee torque). For disassembly of the 400kg, 1.2m by 2m Hyundai Ioniq 5 battery pack in a human-oriented or shared-autonomy work environment such as in Fig. ~\ref{fig:teaser}, full-scale humanoids like the Unitree H1-2 with larger workspaces and torque envelopes (1.78\,m, 70\,kg, 360\,Nm) are required. 

\begin{figure*}[t]
    \centering
    \includegraphics[width=\textwidth]{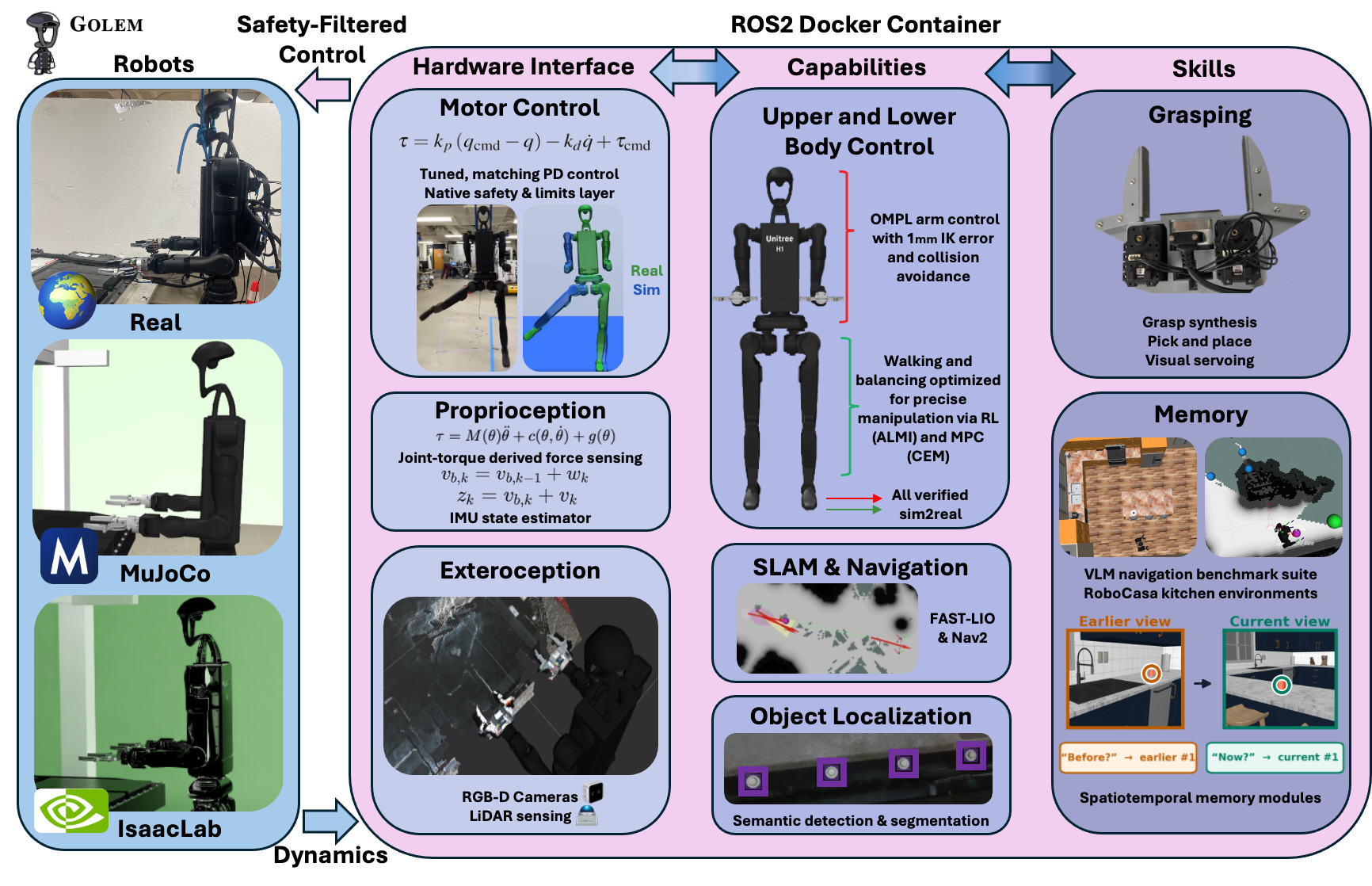}
    \vspace{-20pt}
    \caption{\GOLEM~is an open-source ROS2 development platform for industrial humanoid
    robots, targeting real-world EV battery disassembly. Four interoperable layers connect
    \textbf{(1)}~ Robot interfaces engines (real, MuJoCo, IsaacLab/PhysX) through
    \textbf{(2)}~hardware interfaces for motor control and perception to
    \textbf{(3)}~capabilities spanning learned \& model-based control, navigation, and object
    localization, and \textbf{(4)}~skills for precise manipulation and spatial memory.
    Evaluation spans RoboCasa kitchen \cite{nasiriany2024robocasa} and Hyundai Ioniq~5
    battery environments.}
    \label{fig:system}
    \vspace{-20pt}
\end{figure*}

We present \GOLEM~(Generalized Open Library of Embodied Modules), an open-source, end-to-end system architecture for the Unitree H1-2 humanoid evaluated on simulated RoboCasa kitchen \cite{nasiriany2024robocasa} and sim-to-real EV-LIB disassembly tasks. Within \GOLEM, walking, manipulation, dynamic stability, and navigation are independent modules which can connect to arbitrary control policies. 
\GOLEM~also provides digital twins in MuJoCo \cite{todorov2012mujoco, nasiriany2024robocasa} and IsaacLab \cite{mittal2025isaaclab}, both isolated and communicating identically to the real robot via a Linux and MacOS-compatible ROS2 Docker container, thereby affording flexible switching and evaluation between various methods to compose and verify task-appropriate humanoid autonomy.

The contributions of this paper are twofold:
\vspace{-1pt}
\begin{enumerate}
\item We demonstrate humanoid fastener removal from a Hyundai Ioniq 5 battery pack at three levels of autonomy: tethered, standing, and standing with pose variance post-navigation, characterizing module interplay and where simulation fidelity fails to predict real-world performance.
\item We introduce \GOLEM, an open-source, modular humanoid development platform wherein capabilities are made interchangable on matching simulated \& real interfaces, enabling fair comparison of arbitrary methods.
\end{enumerate}
Finally, we perform a case study implementing multiple Language-Conditioned Spatial Memory methods and use \GOLEM's modular structure and common interfaces as a scaffold to orchestrate our comparison.

\section{Related Work}

\subsection{Robotic Disassembly of EV Batteries}
Disassembling EV battery packs enables material recovery \cite{harper2019recycling} but remains largely manual; robotic work targets the smaller class of hybrid packs \cite{qu2024robotic,wegener2015robot} or task planning \cite{choux2021task}. RAPID \cite{allen2026robotic} addresses full-size 800V packs and solves the unscrewing task (fastener operations account for about 75\% of manual disassembly labor). Then, fastener removal, also known as peg-hole separation \cite{su2023design} or peg-hole disassembly \cite{liu2025integration}, is a difficult task that requires precise control and benefits from compliance. While RAPID can be extended to also remove screws, such a system requires disassembly factories to be designed around it. A humanoid is complementary to the RAPID system and infrastructure in facilities designed for human workers, and we demonstrate \GOLEM~removing screws from a commercial EV pack.

\subsection{Learned Whole-Body Control and Loco-Manipulation}
Reinforcement learning and large-scale motion tracking have produced increasingly capable whole-body controllers. OmniH2O \cite{he2024omnih2o} and HumanPlus \cite{fu2024humanplus} learn whole-body policies from human motion, ALMI \cite{shi2025almi} adversarially couples a lower-body locomotion policy with an upper-body motion-tracking policy on the full-size Unitree H1-2, SONIC \cite{luo2025sonic} scales motion tracking into a generalist controller for the Unitree G1, and LEGS \cite{kim2026legs} leverages photo-realistic gaussian splatting to transfer loco-manipulation policies trained in simulation to a real Unitree G1. 
On the H1-2, Humanoid-COA \cite{wen2025humanoidagentembodiedchainofaction} and Calvert \cite{calvert2026fastresilientadaptablelocomanipulation} demonstrate compositional loco-manipulation behaviors on the Unitree H1-2 with embodied LLM reasoning and behavior trees. 
Unlike a single method or black-box policy, ~\GOLEM~enables learned, model-based, and other various controllers to be wrapped as interchangeable modules and compared against their counterparts. We demonstrate this with MuJoCo-MPC \cite{howell2022predictive} and ALMI\cite{shi2025almi} in Sec.~\ref{sec:experiments}.

\subsection{Middleware, Simulation, and Modular Software}
ROS~2 provides communication, hardware abstraction, and reusable packages for modular robot software \cite{macenski2022robot}; MuJoCo enables physics-based evaluation and model-based control \cite{todorov2012mujoco, howell2022predictive}; RoboCasa provides large-scale simulated environments for everyday tasks\cite{nasiriany2024robocasa}; and IsaacLab facilitates high-fidelity synthetic sensor data generation in GPU-accelerated environments \cite{mittal2025isaaclab}. 
Currently individual capabilities are built upon specialized tools: Open Motion Planning Library (OMPL) for upper-body motion planning \cite{guo2026ompl2}, Nav2 for navigation \cite{macenski2020marathon}, Pinocchio for rigid-body dynamics \cite{carpentier2019pinocchio}. Typically users must hand-craft and assemble these tools and capabilities into a complete control stack; \GOLEM~reduces engineering labor via its shared abstract interfaces and accelerates system-level comparison with an end-to-end architecture, as well as tool \& capability addition. Finally, \GOLEM~empowers developers and broad collaboration with accessible MuJoCo and IsaacLab digital twins which expose ROS~2/DDS interfaces matching the physical robot. 


\vspace{-4pt}
\section{System Architecture}
\vspace{-2pt}

Figure~\ref{fig:system} gives an overview of \GOLEM. We describe its four layers in turn: the container-based robots (Sec.~\ref{sec:physics}), the hardware interfaces for motor control and perception (Sec.~\ref{sec:Hardware}), the core capabilities of control, navigation, and detection (Sec.~\ref{sec:Capabilities}), and the grasping and memory
skills (Sec.~\ref{sec:skills}). Throughout, $q \in \mathbb{R}^{d=27}$ denotes the robot's joint configuration, $a^l \in \mathbb{R}^{d_l=12}$ and $a^u \in \mathbb{R}^{d_u=15}$ the commanded lower- and upper-body joint positions. 
\vspace{-4pt}
\subsection{Container-based robot architecture}\label{sec:physics}
\GOLEM~is implemented as a set of Docker containers that communicate over a single ROS~2 domain with CycloneDDS. A simulation container optionally runs the digital twin (MuJoCo or IsaacLab), which publishes matching interfaces of the physical robot. Currently for IsaacLab, \GOLEM~provides verified support only for hardware interfaces (Sec. ~\ref{sec:Hardware}).  

A second container holds the ROS~2 workspace comprising a bringup that launches locomotion, differential inverse kinematics, perception servers, SLAM, navigation, safety monitoring, and the skill servers that expose capabilities such as grasping and exploration as ROS~2 actions.

 Since the simulator reproduces the real robot's interfaces exactly, switching between simulation and hardware amounts to selecting the DDS domain.
 We emphasize that the twin's value lies in this interface equivalence rather than in dynamic fidelity: contact-rich interactions such as screw engagement are not faithfully reproduced in simulation, which we observe in our sim-to-real experiments.

\subsection{Hardware Interfaces}\label{sec:Hardware}
\textit{Safety}: All motion commands $a^l, a^u$ pass through a low-level safety controller that runs independently of the control modules. The controller continuously monitors joint positions, velocities, and torques against configured limits and immediately stops the robot when any limit is exceeded, for example if a motor experiences an unexpected torque or a state estimate becomes stale. As limits, we are using 90\% of the maximum values specified by Unitree.

In addition, the safety controller constantly polls an Arduino Uno microcontroller connected to a physical emergency-stop button, so a human supervisor can halt the robot at any time. Loss of communication with the Arduino is itself treated as a stop condition. Since the safety controller subscribes to the same interfaces in simulation and on hardware, safety behavior can be exercised and tested in the digital twin before deployment.

\textit{Motor Control}:
Each motor controller commands the desired joint position $q_{\mathrm{cmd}}$ 
and the gravity-compensation torque $\tau_g$
to the underlying PD servo loops running at $500 \, \mathrm{Hz}$ in the motor firmware.
The desired joint velocity is fixed at $\dot{q}_{\mathrm{cmd}} = 0$, causing the derivative term to act as velocity damping that suppresses oscillation.
The firmware therefore computes the actuator torque as
\begin{equation}
    \tau = k_p\left(q_{\mathrm{cmd}} - q\right) - k_d\dot{q} + \tau_{\mathrm{cmd}}
\end{equation}
with $\tau_{\mathrm{cmd}}=\tau_g$.
To reject steady-state position errors caused by model inaccuracies and unmodeled loads, the high-level controller maintains an additional integral bias torque $\tau_b$.
At each time step, the bias torque is updated according to
\begin{equation}
    \tau_b \leftarrow \tau_b + k_i\left(q_{\mathrm{cmd}} - q\right)\Delta t.
\end{equation}
The bias torque is added to the gravity-compensation term, yielding the total commanded  torque
$\tau_{\mathrm{cmd}} = \tau_g + \tau_b.$


\textit{MAGPIE Hands}:
The H1-2 is equipped with parallel jaw MAGPIE \cite{correll2024versatile} hands, a low-cost, force-controlled parallel gripper with built-in 3D perception \cite{correll2024versatile}. Two independently torque-controlled servomotors (Dynamixel AX-12A) drive the fingers through four-bar linkages, providing grasp forces of up to 32N controllable in increments of 0.08N, contact detection, and compliant off-center grasping. This force-controlled architecture also supports integration with semantic physical reasoning frameworks like DeliGrasp \cite{xie2024deligraspinferringobjectproperties} to modulate applied grasp forces based on LLM-inferred material properties. A palm-mounted Intel RealSense D405 observes the object, minimizing occlusion compared to head-mounted cameras and improving fine-grained resolution compared to wrist-mounted cameras. The gripper weighs 414\,g, is built from 3D-printed and commercial off-the-shelf parts for approximately \$460, and its hardware and software are open source. The MAGPIE hands replace the Inspire RH56DFX six-DoF dexterous hands typically paired with the H1-2, since they require significant reorientation and translation to pinch in-place, causing instability \cite{tan2026characterizationanalyticalplanninghybrid}.

\subsubsection{Proprioception}\label{sec:proprioception}
The Unitree H1-2's 27 joints are actuated by high-torque, quasi-direct-drive motors, enabling direct external force observation through FOC current sensing. The lower-body knee, hip, torso, and ankle joints have a torque limit of 360, 220, 220, and $75 \, \mathrm{Nm}$, respectively, and the upper-body shoulder, elbow, and wrist joints have a limit of 120, 120, and $30 \, \mathrm{Nm}$, respectively \cite{unitreeh12motors}.

\textit{Base State Estimation}:
Without ground truth center-of-mass (CoM) measurements we must infer quantities like CoM velocity from the IMU orientation $\bm{R}_b$, angular rate $\bm{\omega}_b$, and joint encoders $\bm{q},\dot{\bm{q}}$; so we track the base velocity $\bm{v}_b\in\mathbb{R}^3$ with a random-walk Kalman filter,
\begin{equation}
\bm{v}_{b,k} = \bm{v}_{b,k-1} + \bm{w}_k,
\qquad
\bm{z}_k = \bm{v}_{b,k} + \bm{v}_k,
\label{eq:rwkf}
\end{equation}
whose measurement is the \emph{planted-foot constraint}~\cite{bloesch2012}: in which each foot contact $k$ gives 
\begin{equation}
\bm{z}_k = -\Big[\, \bm{\omega}_b \times \big(\bm{p}_k(\bm{q}) - \bm{p}_b\big)
    + \bm{R}_b\,\bm{J}_k(\bm{q})\,\dot{\bm{q}} \,\Big],
\label{eq:legodom}
\end{equation}
with $\bm{J}_k$ the translational Jacobian of contact $k$ (one forward-kinematics evaluation). Feet are fused with a standard Kalman update using contact-weighted, outlier-gated measurement noise~\cite{camurri2017}; orientation comes from the IMU, whereas horizontal position, unobservable from proprioception~\cite{bloesch2012,rotella2014}, is integrated open-loop.

\subsubsection{Exteroception}
The exteroceptive suite is designed to handle navigation, approach, and precise manipulation at variable distances. For navigation, SLAM and obstacle avoidance are handled by a head-mounted Livox MID-360 LiDAR. On the approach, mid-range object localization and VLM perception rely on the robot's stock forward-facing Intel RealSense D435i camera. Then, for precise manipulation, hand-mounted RealSense D405 cameras, specialized for sensing between $7 \, \mathrm{cm}$ and $50 \, \mathrm{cm}$, facilitate closed-loop visual servoing to reduce accumulated forward-kinematic and head camera calibration errors.

\subsection{Capabilities}\label{sec:Capabilities}

\subsubsection{Lower-Body Control}
\label{sec:lowerbody}

This module exposes a common interface for the H1-2's 12 leg joints, mapping commanded base twists $u_{\mathrm{base}} = (v_x, v_y, \omega_z)$ to lower body commands $a^l$ for the servo layer. The waist and arms remain controlled by the IK stack (Sec.~\ref{sec:upperbody}). Input twists are clamped to the active controller's valid envelope, and standing is handled implicitly as a zero-twist command. We integrate three controllers, which are wrapped in the \GOLEM~ interface for fair comparison:

\begin{enumerate}
    \item \textbf{MJPC~\cite{howell2022predictive}} A sampling-based model predictive controller planning at $67 \, \mathrm{Hz}$ with a 1s planning horizon.
    \item \textbf{ALMI~\cite{shi2025almi}:} A learned policy running at $50 \, \mathrm{Hz}$. Stance behavior emerges implicitly from reward terms gated on $\|u_{base}\| < 0.1$.
    \item \textbf{Vendor Policy:} Unitree's default black-box controller.
\end{enumerate}

\textit{Model-Based Lower-Body Control}:
We formulate lower-body locomotion and dynamic stabilization as a finite-horizon optimal control problem solved in receding-horizon fashion using MuJoCo-MPC (MJPC) \cite{howell2022predictive}. Let $x_t = (q_t, \dot q_t, b_v)$ denote the whole-body state and base velocity;  at each control time step, the planner solves
\begin{equation}
\min_{a^l_{0:T}} \; \sum_{t=0}^{T} \ell(x_t, a^l_t)
\quad \text{s.t.} \quad x_{t+1} = f(x_t, a^l_t, \Delta t),
\label{eq:ocp}
\end{equation}
where $f$ is MuJoCo's contact dynamics \cite{todorov2012mujoco} and $\ell$ is a weighted sum of task-residual norms over objectives such as center-of-mass height and velocity, torso orientation, foot placement, and control effort. In MJPC, we employ predictive sampling (CEM) \cite{deboer2005tutorial} to optimize Eq. \eqref{eq:ocp} in real time, warm-starting each step by the time-shifted plan. 

\textit{Learning-Based Lower-Body Control (ALMI):}
\label{sec:almi}
ALMI~\cite{shi2025almi} formulates locomotion training as a two-player zero-sum Markov game between the lower-body policy $\pi^l$ and an adversarial upper-body policy $\pi^u$ sharing state $s_t$ comprised of proprioception, commanded velocity, and phase parameter. To ensure robust velocity tracking under dynamic upper-body disturbances, $\pi^l$ maximizes the command-following return $r^l$ while $\pi^u$ acts as an adversary seeking to minimize it:
\begin{equation}
\max_{\pi^l} \min_{\pi^u} \mathbb{E}\left[ \sum_{t=0}^{T} r^l(s_t, a^l_t, a^u_t) \right]
\end{equation}

 During locomotion training, $\pi^u$ generates adversarial actions $a^u_{\mathrm{adv}} \sim \pi^u(\cdot \mid s_t^{\mathrm{prop}}, g^u)$ driven by reference posture commands $g^u$ sampled via a dual-difficulty curriculum. This min-max formulation hardens $\pi^l$ against dynamic recoil, payload swings, and momentum shifts. In \GOLEM, we instantiate only the 12-DoF leg policy $\pi^l$ through the lower-body control interface, and replace $\pi^u$ with our upper-body controller (Sec.~\ref{sec:upperbody}). 

\subsubsection{Upper-Body Control}\label{sec:upperbody}
Upper-body motion is driven by desired end-effector poses. The upper-body controller interface expects two end-effector poses in $SE(3)$ and outputs upper-body motor commands $a^u$ to achieve these poses. 
Upper-body motion generation combines differential inverse kinematics with sampling-based motion planning.
\GOLEM~ integrates direct differential inverse kinematics for short, unobstructed motions and sampling-based planning for larger motions that require collision avoidance.
For efficient collision checking, \GOLEM~ uses MorphIt \cite{nechyporenko2026morphit} to construct a packed-sphere approximation of the robot’s upper body,
reducing the median collision-checking time from
$38.8\ \mu\mathrm{s}$ with the mesh representation
to
$5.8\ \mu\mathrm{s}$
($6.7\times$ speedup).

\textit{Differential Inverse Kinematics}:
End-effector goals are tracked by a task-space controller implemented with Pink \cite{caron2024pink}. Each task $i$ defines a residual $r_i(q)$, e.g., the error between the current and target end-effector pose, whose Jacobian $J_i(q) = \partial r_i / \partial q$ is computed with Pinocchio \cite{carpentier2019pinocchio}. At each control step, the controller solves the quadratic program
\begin{equation}
\min_{\dot q} \; \sum_i w_i \big\| J_i(q)\,\dot q - \alpha_i r_i(q) \big\|^2 + \lambda \|\dot q\|^2
\;\; \text{s.t.} \;\; \dot q^- \le \dot q \le \dot q^+,
\label{eq:qp}
\end{equation}
where the $w_i$ weight competing tasks, $\alpha_i$ are task gains, $\lambda$ regularizes the solution, and the box constraints enforce joint position and velocity limits. The optimizer $\dot q^\star$ is integrated as $q \leftarrow q + \dot q^\star \Delta t$ to produce $a^u$.

\textit{Sampling-Based Planning}:
For longer-range motions in cluttered environments, we use OMPL's implementation of RRT-Connect \cite{sucan2012ompl} to generate collision-free paths in the 15-DOF upper-body joint space. The sampled configurations are validated
against joint limits, self-collision, and configurable workspace constraints. For battery disassembly, the position of each grasp frame is constrained to a fixed Cartesian region in front of the robot to limit disturbances to the balance:
\begin{equation}
x \leq 0.60\,\mathrm{m}, \ \
y \in [-0.45,\,0.45]\,\mathrm{m}, \ \
z \in [-0.05,\,0.60]\,\mathrm{m}.
\end{equation}

\subsubsection{Object Localization}\label{sec:object-localization}
\GOLEM's modular architecture easily incorporates any bounding box or segmentation mask producing model. The \GOLEM~interface expects object detection models to output point clouds of requested objects from RGB-D cameras. In generic scenes like RoboCasa we leverage internet-scale VLMs like Gemini Robotics \cite{geminiroboticsteam2025geminiroboticsbringingai}, and for niche battery components, we fine-tune YOLO-World \cite{yoloworld} on battery components such as screws, nuts, and bus bars. RGB detections are applied to the RGB-D point cloud, where they can be consumed by higher-level skills.
\subsubsection{Navigation}
Navigation builds on a ROS~2 \& Nav~2 stack \cite{macenski2020marathon}. FAST-LIO \cite{xu2022fast} provides LiDAR-inertial odometry and mapping from the robot's Livox MID-360; the resulting point cloud is projected into a 2D occupancy grid, while the full 3D cloud is incorporated into the costmap for collision checking. Nav2's global path planner performs a graph search over this costmap: writing $C : \mathcal{G} \to [0, c_{\max}]$ for the cost of traversing grid cell $g \in \mathcal{G}$, it finds a cell path $g_{0:N}$ from the robot's cell to the goal cell minimizing
\begin{equation}
\sum_{k=0}^{N-1} \Big( \left\| g_{k+1} - g_k \right\| + \beta\, C(g_{k+1}) \Big),
\label{eq:nav}
\end{equation}
using A$^{*}$ search with an admissible Euclidean heuristic, where $\beta$ trades path length against clearance from obstacles.
A local controller tracks the resulting path and emits velocity commands to the general lower-body controller interface.

\subsection{Skills}\label{sec:skills}
Skills are the highest level of abstraction in \GOLEM; skills orchestrate capabilities through their well-defined interfaces and enable completion of high level objectives.

\textit{Grasping}:
To evaluate grasping within the \GOLEM~architecture, we implement manipulation modules with a standardized skill interface. We compare multiple grasp synthesis strategies within the same architecture: (i) an open-world pipeline in which an object is detected via the object localization capability ~\ref{sec:object-localization}, from which the segmented object's point-cloud can be passed as input to neural six-DoF grasp pose synthesis methods such as GraspGenX (GGX) \cite{graspgenx2026} or classical methods such as antipodal grasp sampling; and (ii) a top-down visual-servoing approach that aligns the palm-mounted or otherwise wrist-mounted camera with the target before closure.

\textit{Spatiotemporal Memory}: 
Spatiotemporal memory expands ~\GOLEM's spatial understanding beyond the robot's current field
of view \cite{li2026spatial}. An agent can query an object or region that is not presently visible,
navigate to the returned location in memory, and then use local perception and manipulation
skills.
\GOLEM~ uses VLMaps~\cite{huang23vlmaps} to back-project dense visual-language features from posed RGB-D observations and fuse them into a persistent top-down semantic map. Following the retrieval-augmented memory architecture, ReMEmbR~\cite{anwar2025remembr}, we also include a backend adapted to store SigLIP image embeddings together with their poses and timestamps. An optional VLM re-ranks the retrieved candidates based on the query and their temporal metadata.

During memory retrieval, \GOLEM~accepts a free-text query referring to an object and returns candidate locations, relevance scores, and backend metadata. Exploration, navigation, and agentic programs can therefore access spatial and temporal information through a common interface, independent of the underlying memory implementation.

\begin{figure}[!t]
    \centering
    \includegraphics[width=0.99\linewidth]{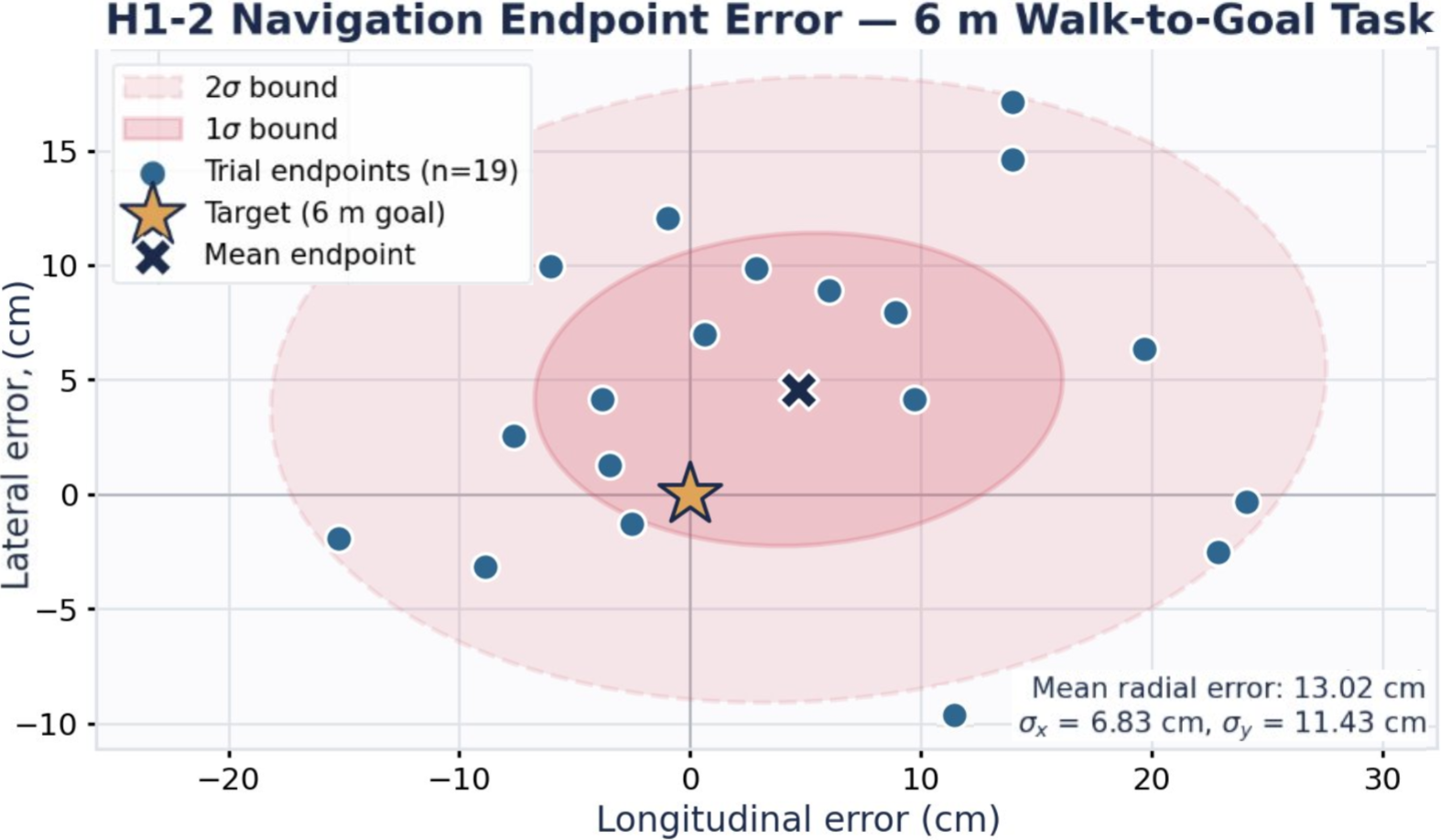}
    \caption{Six meter start-to-goal walking accuracy using the FAST-LIO and Nav2 navigation stack to localize from noisy initialization with an off-the-shelf Unitree walking policy.}
    \label{fig:navigation}
    \vspace{-12pt}
\end{figure}

\vspace{-4pt}
\section{Experiments}\label{sec:experiments}
\vspace{-4pt}
\begin{figure*}[t]
    \centering
    \includegraphics[width=0.99\textwidth]{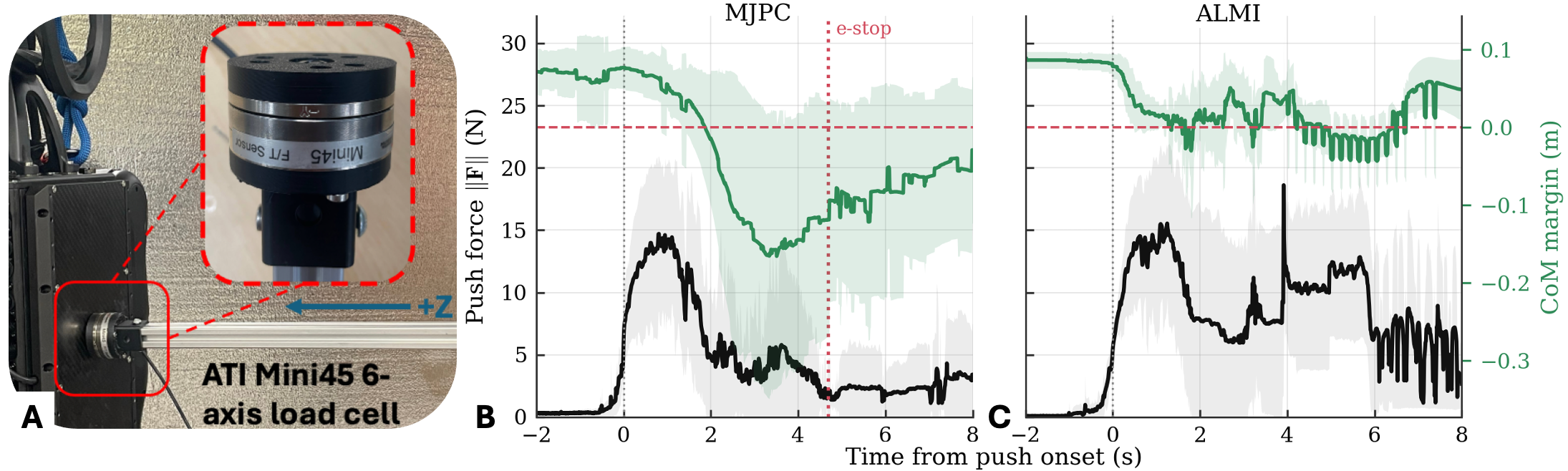}
    \caption{\textbf{Standing with force adaptivity, MPC vs. RL} We evaluate standing robustness with \textbf{(a)~}a pushing rod mounted with a ATI Mini45 six-axis load cell applying an oppositional force to the robot torso, \textbf{(b-c)~}indicated in \textbf{black} until the robot loses balance, measured by the center-of-mass (CoM) margin from the support polygon (\textcolor{green}{green}) becoming negative (horizontal \textcolor{red}{red} line). Both standing methods exhibit similar initial force adaptivity, but \textbf{(b)~} sampling-based model predictive control (MJPC) is not able to recover from the initial loss in balance and must be emergency stopped (vertical \textcolor{red}{red} line). In comparison, \textbf{(c)~}adversarial reinforcement learning (ALMI) exhibits a recovery behavior to restore a positive CoM margin.}
    \label{fig:pokey_data}
    \vspace{-15pt}
\end{figure*}
We structure the evaluation as a capability ladder, enabled by \GOLEM: each experiment introduces one additional module, quantifies the error it contributes, and passes that error on to the next. We first establish navigation accuracy under interchangeable walking policies (Sec.~\ref{sec:exp-navigation}), then the robot's ability to hold position against unmodeled forces (Sec.~\ref{sec:push}), and finally we evaluate grasping of Hyundai Ioniq 5 battery pack screws at three levels of autonomy that successively integrate these modules (Sec.~\ref{sec:exp-grasping}).

\subsection{Walking Policy and Navigation Accuracy}
\label{sec:exp-navigation}
The Unitree-provided H1-2 walking policy wrapped in \GOLEM's velocity interface subscribes to \GOLEM's FAST-LIO and Nav2 stack and is evaluated on a 6\,m straight-line point-to-point navigation task over 20 trials, reaching the goal within a mean radial error of 13.0\,cm ($\sigma_x{=}6.83$\,cm, $\sigma_y{=}11.43$\,cm) on 19 of 20 runs (95\%), shown in Fig. ~\ref{fig:navigation}, with the failed trial veering several meters off due to policy instability.
This endpoint error defines the base-placement variance that downstream manipulation must tolerate when the robot walks to its work station (Sec.~\ref{sec:exp-grasping}). Preliminary testing of ALMI for walking demonstrated significant inconsistency in stopping and stabilizing to a standing stance, precluding it from further evaluation.

\subsection{Force-Adaptivity}\label{sec:push}
We compare the force-adaptivity of two standing policies, adversarial RL-based ALMI and the lower-body MPC. Preliminary testing of the vendor-provided policy showed an inability to stand still, precluding it from standing evaluation and motivating policy switch-over to a distinct standing policy.
Pushes are applied manually at the pelvis with a load-cell mounted instrumented rod
(Fig.~\ref{fig:pokey_data}). Mean peak per-trial force is similar between MJPC ($20.4 \pm 7.0\,$N) and ALMI
($22.1 \pm 6.9\,$N), and in both controllers the CoM is driven outside the support polygon by the push. However, MJPC decreases monotonically to a $-0.166\,$m margin with
$16.3 \pm 6.3^\circ$ peak pitch-axis lean, requiring an operator e-stop at $4.68 \pm 2.86\,$s. In contrast, ALMI reaches a minimum
margin of $-0.044$~m with $6.8 \pm 1.4^\circ$ peak pitch-axis lean and notably recovers a positive
margin in 93\% of trials (14/15). Due to this desirable recovery behavior and robustness, subsequent experiments deploy ALMI for lower-body standing. 

\newcommand{\wc}[3]{\shortstack{#1\\[-1.5pt]{\tiny[#2,\,#3]}}}

\begin{figure*}[t]
  \centering
  \begin{subfigure}[b]{0.33\linewidth}
    \centering
    \setlength{\tabcolsep}{2.5pt}%
    \renewcommand{\arraystretch}{1.15}%
    \scriptsize
    \begin{tabular}{@{}l ccccc@{}}
      \toprule
      \textit{Task} (Env) & Cent. & Antip. & GGX & GGX-F & C-VS \\
      \midrule
      \multicolumn{6}{@{}l}{\emph{Fridge} (sim)} \\
      \quad Tether & \textbf{\wc{100}{89}{100}} & \wc{63.3}{46}{78} & \wc{53.3}{36}{70} & \wc{96.7}{83}{99} & \na \\
      \quad Stand  & \textbf{\wc{80.0}{63}{90}}   & \wc{26.7}{14}{44} & \wc{53.3}{36}{70} & \wc{70.0}{52}{83} & \na \\
      \multicolumn{6}{@{}l}{\emph{Screw} (sim)} \\
      \quad Tether & \textbf{\wc{100}{89}{100}} & \na & \wc{0}{0}{11} & \wc{0}{0}{11} & \textbf{\wc{100}{89}{100}} \\
      \quad Stand  & \wc{90.0}{74}{97} & \na & \wc{0}{0}{11}  & \wc{0}{0}{11} & \textbf{\wc{100}{89}{100}} \\
      \multicolumn{6}{@{}l}{\emph{Screw} (real)} \\
      \quad Tether & \wc{0}{0}{11} & \na & \na & \na & \textbf{\wc{96.7}{83}{99}} \\
      \quad Stand  & \wc{0}{0}{11} & \na & \na & \na & \textbf{\wc{86.7}{70}{95}} \\
      \quad Var. Stand & \wc{0}{0}{11} & \na & \na & \na & \textbf{\wc{36.7}{22}{55}}\\
      \bottomrule
    \end{tabular}
    \caption{%
      Grasp success (\%, 95\% confidence intervals below) across two tasks, three levels of autonomy, and five methods ($n=30$ trials).}
    \label{fig:sub-table}
  \end{subfigure}%
  \begin{subfigure}[b]{0.66\linewidth}
    \centering
    \includegraphics[width=\linewidth]{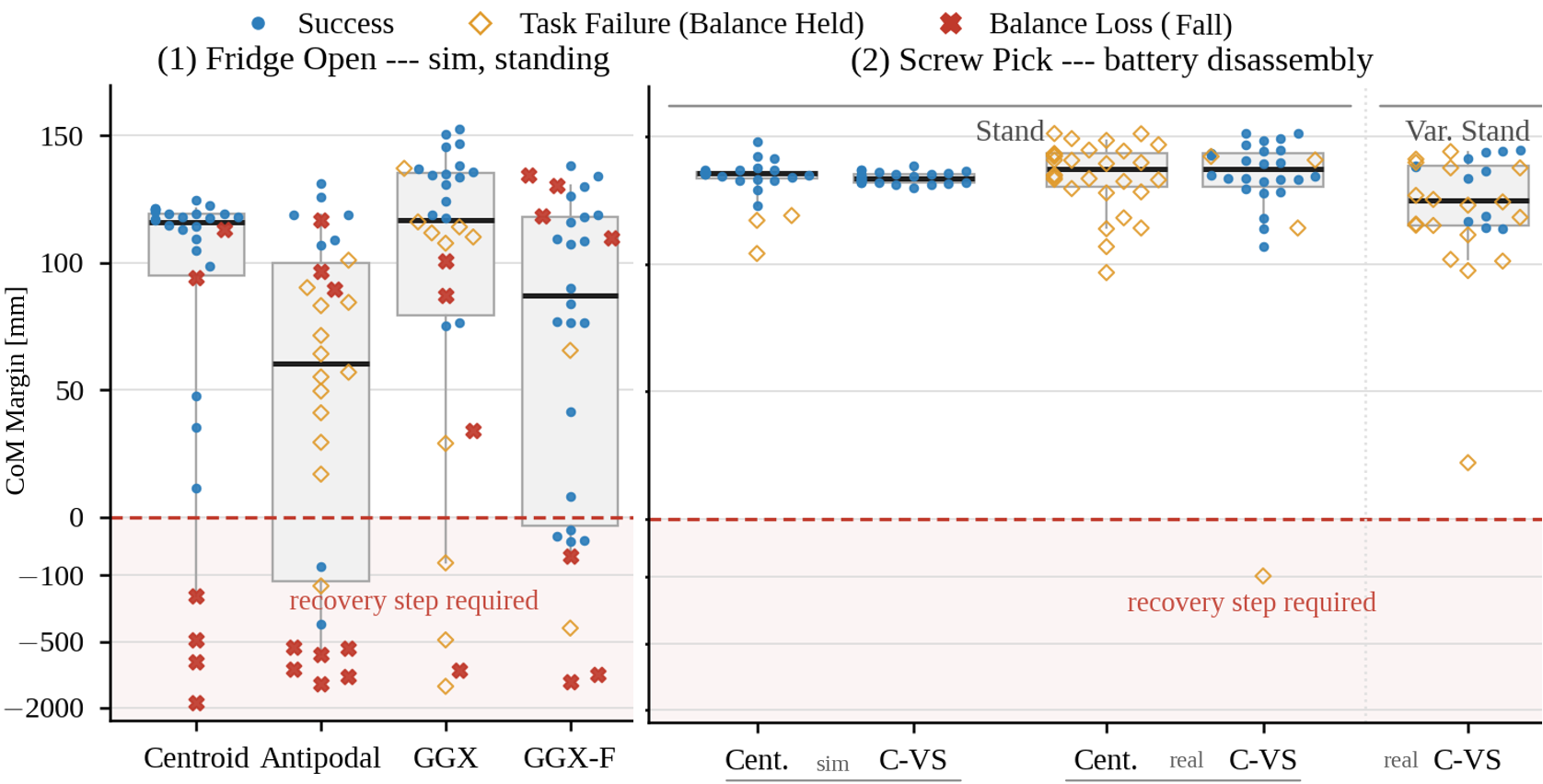}
    \caption{center-of-mass (CoM) margin evaluation across simulated and real-world trials.}
    \label{fig:sub-image}
  \end{subfigure}
  \caption{\textbf{(a)} Grasp success on fridge-handle opening and Ioniq~5 screw
    picking, at three autonomy levels: \emph{Tether} (hoist-assisted), \emph{Stand} (ALMI
    balancing), \emph{Var.\ Stand} (base pose from the navigation placement distribution).
    Methods: centroid (Cent.), antipodal (Antip.), GraspGenX \cite{graspgenx2026} (GGX),
    reachability-reranked GGX (GGX-F), centroid visual servoing (C-VS). \textbf{(b)}
    Episode-level average CoM margin; negative values (\textcolor{red}{red}) indicate a
    recovery step or fall. Boxes are median and IQR, 10--90\,\% whiskers.}
  \label{fig:combined-results}
  \vspace{-20pt}
\end{figure*}

\subsection{Manipulation}\label{sec:exp-grasping} We evaluate the five following grasp synthesis baselines (Fig. ~\ref{fig:sub-table}):
Centroid grasps the segmented object centroid with a fixed top-down or forward
approach; Antipodal samples antipodal pairs from the point cloud; GGX samples
grasps from GraspGenX \cite{graspgenx2026}; GGX-F reranks GGX candidates by wrist reachability;
C-VS closes the loop on the centroid with visual servoing.
We select two task domains: grasping and opening of a fridge handle in RoboCasa kitchens and grasping of screws loosened by the RAPID gantry \cite{allen2026robotic} from the Hyundai Ioniq 5 pack. Finally, we evaluate our methods on three autonomy levels that successively integrate the modules validated above: (a) \emph{tethered}, with the robot supported by a hoist so that its legs are unloaded, isolating the perception and manipulation modules from balance; (b) \emph{standing}, with the robot balancing at a pre-computed screw-reachable base pose, adding the standing controller of Sec.~\ref{sec:push}; and (c) \emph{standing with pose variance}, with the robot first navigating to the pack using the stack of Sec.~\ref{sec:exp-navigation}, adding its base-placement error to the grasping problem. The difference in success rate between consecutive conditions isolates the contribution of each module to overall task failure. In Fig. ~\ref{fig:sub-image}, stability sensitivity to grasp methods is measured by center-of-mass XY-projected signed distance to the H1-2 support polygon boundary  (CoM margin).

\textit{Fridge opening (sim).}
Under tether, centroid (100\%) and GGX-F (96.7\%) methods are effective, while
antipodal (63.3\,\%) and GGX (53.3\,\%) are less reliable. Standing incurs
a further 20.0, 26.7, 0.0, and 36.6\% success reduction, respectively (Fig.~\ref{fig:sub-table}). Due to centroid grasping generating a single, stereotyped grasp in each trial, it is the most stable method with a CoM margin distribution (median 115.8\,mm, IQR
95.0--119.3\,mm), $2.3$--$9.0\times$ narrower than any pose-sampling method,
and only 13.3\,\% of its episodes push the projected CoM outside the support
polygon, against 26.7\,\% (GGX-F) and 30.0\,\% (antipodal). 

\textit{Screw picking (sim).}
On the screw, both GGX and GGX-F fail completely at both tiers due to imprecision in grasp generation, while centroid (90.0\,\%) and C-VS
(100\,\%) are effective and do not leave the support polygon.

\begin{table*}[!b] \centering \caption{Comparison of memory modules through \GOLEM's common interface. Average success over 20 RoboCasa episodes; bounds show 95\% confidence intervals.} \label{tab:spatial_memory} \scriptsize \setlength{\tabcolsep}{4pt} \renewcommand{\arraystretch}{1.25} \begin{tabular}{lccccc} \hline Method & \shortstack{Basic retrieval:\\Static MRR \(\uparrow\)} & \shortstack{Temporal awareness:\\Current-location Top-1 \(\uparrow\)} & \shortstack{Temporal awareness:\\Outdated-location Top-1 \(\downarrow\)} & \shortstack{Temporal memory:\\History coverage@3 \(\uparrow\)} & \shortstack{Efficiency:\\Median query time \(\downarrow\)} \\ \hline VLMaps & \(77.5_{60.9}^{94.1}\%\) & \(20.0_{2.0}^{38.0}\%\) & \(53.5_{31.6}^{75.3}\%\) & \(45.0_{35.2}^{54.8}\%\) & \(20.4_{19.5}^{21.3}\) ms \\ SigLIP+FAISS & \(67.5_{50.2}^{84.8}\%\) & \(35.7_{15.7}^{55.8}\%\) & \(41.4_{19.7}^{63.1}\%\) & \(42.5_{26.2}^{58.8}\%\) & \(3.6_{3.5}^{3.6}\) ms \\ SigLIP+FAISS+rerank & \(100.0\%\) & \(95.4_{92.2}^{98.6}\%\) & \(0.0\%\) & \(97.5_{92.6}^{100}\%\) & \(\approx 19.7_{18.4}^{21.0}\) s \\ \hline \end{tabular} \vspace{-15pt} \end{table*}

\textit{Screw picking (real).}
Reflecting a large sim-to-real gap in visual uncertainty, open-loop centroid fails completely in real trials despite CoM margins matching simulation (median 137.2 vs.\ 133.5\,mm). Imperfect depth-segmentation and
head camera calibration produce significant positional error ($\geq$5cm) that control from single-shot detection is not robust to. Visual servoing recovers
near-perfect performance (96.7\% task success) while tethered, and 86.7\,\% while standing. Base motion sways significantly more in the real-world deployment (radial sway RMS 4.6mm) with a median peak base excursion of
18.0\,mm per reach, compared to sway in simulation (0.10\,mm), which effectively is static.  

Despite this increased sway, in all C-VS trials the initial grasp closes on the screw; the five failures across real-world tether and stand removal are caused by off-axis extraction. Deviation from a vertical extraction due to non-linear OMPL motion plans at robot joint limits cause the screw to jam against the bore wall. Then, the gripper's grasp force is insufficient to overcome the resulting friction, leading to slip. Across all C-VS trials, the average wallclock time from image detection trigger to grasp completion is 75$\pm$16.1s.
\noindent\textit{Placement variance (Var.\ Stand).}
Sampling five grasps with visual servoing at each of six stand poses drawn from the navigation
placement distribution, three near poses achieve 11/15
successful grasps and three (one very near, two far) fail completely since the pose is out of reach. Stability at the near poses (median 137.2mm) decreases to 124.8\,mm ($p=0.022$) at the off-nominal poses. The \GOLEM~architecture enables characterization and pinpointing of failure directly to standing poses, rather than to grasp or removal imprecision.

\subsection{Case Study: Language-Conditioned Spatial Memory}
\GOLEM~ provides a platform for fair comparison of humanoid modules. We demonstrate how heterogeneous memory backends can leverage \GOLEM's common RGB-D, pose, and navigation interfaces for language-conditioned spatial and temporal retrieval benchmarking in Tab. ~\ref{tab:spatial_memory}. The benchmark contains 20 controlled RoboCasa object-relocation episodes: five each for a mug, bowl, can, and bottled water. In each episode, the camera follows the same eight-view route before and after the target moves to a different counter, producing 320 RGB-D observations. The evaluator issues 153 static, current-location, historical-location, and absent-object queries.

Static Mean Reciprocal Rank (MRR) measures basic retrieval quality. Current-location Top-1 measures whether the highest-ranked result reflects the object's present location. Outdated-location Top-1 measures how often stale evidence is ranked first. History coverage@3 measures retrieval of the object's previously observed locations among the top three results. Median query time measures retrieval efficiency. 

SigLIP+FAISS adapted from \cite{anwar2025remembr} provides the lowest retrieval latency but has limited temporal awareness. VLMaps provides map-based spatial aggregation at moderate latency but frequently prioritizes outdated observations after objects move. VLM re-ranking improves retrieval and temporal disambiguation, at the cost of increased query latency on the order of $10^3\times$. \GOLEM~ exposes these alternatives through the same interface, allowing selection of a memory backend suited for a task's spatiotemporal, and runtime requirements.

\vspace{-4pt}
\section{Conclusion}
\vspace{-2pt}
We present \GOLEM, an end-to-end open source system architecture for humanoid autonomy motivated by the industrial task of EV battery disassembly. By exposing simulated and physical resources through identical ROS~2/DDS interfaces and capabilities as reusable modules, the architecture supports development by seamless mixing of the digital twin with real hardware, culminating in a Unitree H1-2 removing screws from a Hyundai Ioniq 5 battery pack.

We have compared grasping and balancing methods for screw removal as part of a Hyundai Ioniq 5 battery pack disassembly process and used \GOLEM~ as a scaffold to perform fair comparisons between functionally equivalent modules. \GOLEM~enables the finding that screw removal errors reside primarily in locomotion error, motivating future work on precise relocation and repositioning. The digital twins provided by \GOLEM~reveal that simulation under-predicts perception error and base-motion coupling, so sim success rates on contact-rich, precise tasks are not predictive.


We also note that as humanoid systems move toward deployment alongside people, the ability to attribute failures to specific components becomes a safety concern rather than only a scientific one; we intend \GOLEM's modular evaluation to support this accountability, while noting that benchmark performance in simulation or controlled settings does not establish readiness for unsupervised operation around humans.
\vspace{-12pt}


\section*{ACKNOWLEDGMENT}
The authors are supported by ARPA-E grant DEAR0001966, ``Robust Robotic Disassembly of EV Battery Packs using Open-World Vision Language Models and Symbolic Replanning.'' William Xie is supported by the NSF
Graduate Research Fellowship. N. Correll and M. Conway have an interest in Realtime Manufacturing, Inc., which is working on humanoids
for manufacturing applications.
\bibliographystyle{ieeetr}
\bibliography{HAMS}

\end{document}